\pdfoutput=1
\documentclass[11pt]{article}

\usepackage{acl}
\usepackage{times}
\usepackage{latexsym}

\usepackage[T1]{fontenc}
\usepackage[utf8]{inputenc}

\usepackage{microtype}
\usepackage[ruled]{algorithm}
\usepackage[noend]{algpseudocode}
\algrenewcommand\algorithmicindent{1em}%
\usepackage{enumitem}% noitemsep
\usepackage{colortbl}
\usepackage{xcolor}
\usepackage{esvect}

\usepackage{url}
\usepackage{booktabs}
\usepackage{scrextend}
\usepackage{graphicx}
\usepackage{textpos}
\usepackage{tikz}
\usepackage[ruled]{algorithm}
\usepackage[noend]{algpseudocode}
\usepackage[normalem]{ulem}
\usepackage{siunitx}
\usepackage{amsmath}
\usepackage{amssymb} % \mathbb
\usepackage{multirow}
\usepackage{lscape}
\usepackage{rotating}
\usepackage{xspace}
\usepackage{subcaption}

\title{TexPrax: A Messaging Application for Ethical, Real-time Data Collection and Annotation}

 \author{Lorenz Stangier\thanks{~~Equal contribution}~~\footnotemark[2] \and Ji-Ung Lee\footnotemark[1]~~\footnotemark[2] \and Yuxi Wang\footnotemark[3] \and Marvin M\"{u}ller\footnotemark[3] \\\textbf{ Nicholas Frick}\footnotemark[3] \and \textbf{Joachim Metternich}\footnotemark[3] \and \textbf{Iryna Gurevych}\footnotemark[2] \\ 
         Ubiquitous Knowledge Processing (UKP) Lab\footnotemark[2] \\ 
         Institute of Production Management, Technology and Machine Tools (PTW)\footnotemark[3] \\
          Department of Computer Science and Hessian Center for AI (hessian.AI)\footnotemark[2] \\
         Department of Engineering\footnotemark[3] \\
         Technical University of Darmstadt}

\begin{document}
\maketitle

\begin{abstract}
Collecting and annotating task-oriented dialog data is difficult, especially for highly specific domains that require expert knowledge.
At the same time, informal communication channels such as instant messengers are increasingly being used at work.
This has led to a lot of work-relevant information that is disseminated through those channels and needs to be post-processed manually by the employees.
To alleviate this problem, we present TexPrax, a messaging system to collect and annotate \textit{problems}, \textit{causes}, and \textit{solutions} that occur in work-related chats.
TexPrax uses a chatbot to directly engage the employees to provide lightweight annotations on their conversation and ease their documentation work. 
To comply with data privacy and security regulations, we use an end-to-end message encryption and give our users full control over their data which has various advantages over conventional annotation tools.
We evaluate TexPrax in a user-study with German factory employees who ask their colleagues for solutions on problems that arise during their daily work.
Overall, we collect 202 task-oriented German dialogues containing 1,027 sentences with sentence-level expert annotations.
Our data analysis also reveals that real-world conversations frequently contain instances with code-switching, varying abbreviations for the same entity, and dialects which NLP systems should be able to handle.\footnote{Code and data are published under an open source license: \url{https://github.com/UKPLab/TexPrax}}
\end{abstract}

%Venue: \url{https://2022.emnlp.org/calls/System_Demonstrations/}

\section{Introduction}\label{sec:introduction}
The lack of annotated data---especially in languages other than English---is one of the key open challenges in task-oriented dialogue processing~\citep{razumovskaia-etal-2022-natural}.
This becomes even more challenging for very task-specific application domains with only a small number of experts that are sufficiently qualified to generate dialogue data or provide annotations~\citep{sambasivan2021everyone}.
At the same time, using informal communication channels such as instant messengers at work has become increasingly popular~\citep{rajendran2019understanding,NEWMAN2021100802}.
Although this can accelerate troubleshooting, most of the knowledge that is communicated informally may be lost without an additional error tracking process; which in turn increases documenting work for employees~\citep{tubiblio125537}.
Whereas this could be alleviated by NLP-based assistance systems---that for instance automatically identify \textit{problems}, their \textit{cause}, and their \textit{solution}---they cannot be built without any annotated data.
Our goal is to provide an application (TexPrax) to bridge the gap between the lack of annotated task-oriented dialogue data and the increasing need for NLP-based documenting assistance.

\begin{figure}
    \centering
    \includegraphics[width=0.48\textwidth]{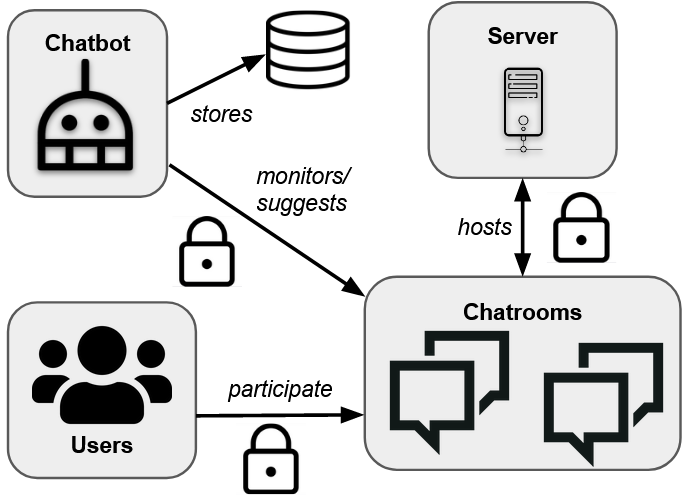}
    \caption{Overview of TexPrax. All users as well as the chatbot communicate via chatrooms that are hosted on a Synapse server instance. All messages are end-to-end encrypted using the Matrix communication protocol.}
    \label{fig:bot_user}
\end{figure}

Figure~\ref{fig:bot_user} provides a high-level overview of TexPrax and all involved parties.
Our system communicates as a chatbot that acts as the user interface and recording service at the same time. 
For the server that hosts the messaging application, the bot appears as an additional user and hence, inherits all privileges and restrictions a user can have; including (1) reading any messages written in a chatroom, (2) being invited and removed by the chatroom moderator, and (3) being able to send messages in chatrooms it was invited to.
We use privilege (3) to provide label suggestions from a pre-trained model and collect annotations via a reaction mechanism (Figure~\ref{fig:msg_classification}); attaining a lightweight annotation process with minimal overhead.
We also integrate TexPrax via the REST web API into an internal dashboard to automatically store recognized errors as a first step of the error documentation process.

Directly involving employees in data annotation and curation introduces four key advantages over previous approaches that involve crowdsourcing~\citep{crowston2012amazon} or use expert annotation tools such as INCEpTION~\citep{klie2018inception}.
First, they are the very domain experts that hold qualified conversations which concern exactly the target-domain. 
This allows us to directly collect the dialog data instead of having to generate it semi-automatically or asking crowdworkers who can only provide limited expertise~\citep{raghu-etal-2021-end}.
Second, the employees have an immediate benefit from annotating and improving the recommendation model as a dashboard integration saves time they would otherwise have to spend on documenting errors later on (hence, an intrinsic motivation).
Third, they have full control over their own data which saves time for NLP practitioners as it alleviates research data management.
Finally, the use of an end-to-end encryption protocol ensures that only parties selected by the employees will have access to the data even if the server is breached.\footnote{Upon creating a chat room, they will explicitly be asked if our chatbot is allowed to join the chatroom (opt-in).}
Our contributions are:

\begin{enumerate}[topsep=5pt,itemsep=2pt]
    \item An application for collecting and annotating dialogues in real-time to assist employees during their work. To comply with data privacy and safety regulations such as the GDPR~\citep{GDPR}, TexPrax further has received full clearance by the ethics committee and staff council of TU Darmstadt.
    \item A German dataset with 202 dialogues, consisting of 591 turns, and 1,027 annotated sentences collected from a highly specific domain, namely an assembly line in a factory.
\end{enumerate}

\section{Use Case}\label{sec:usecase}
In this work, we focus on assisting employees on the shop floor (the production area in a factory).
Our goal is to improve shop floor management~\citep{hertle2017darmstadter}; a systematic approach for solving processing problems.
To efficiently solve such problems, shop floor management defines performance indicators which are used to detect deviations and identify problems which are also used to quantify their successful rectification. 

\begin{figure}
    \centering
    \includegraphics[width=0.48\textwidth]{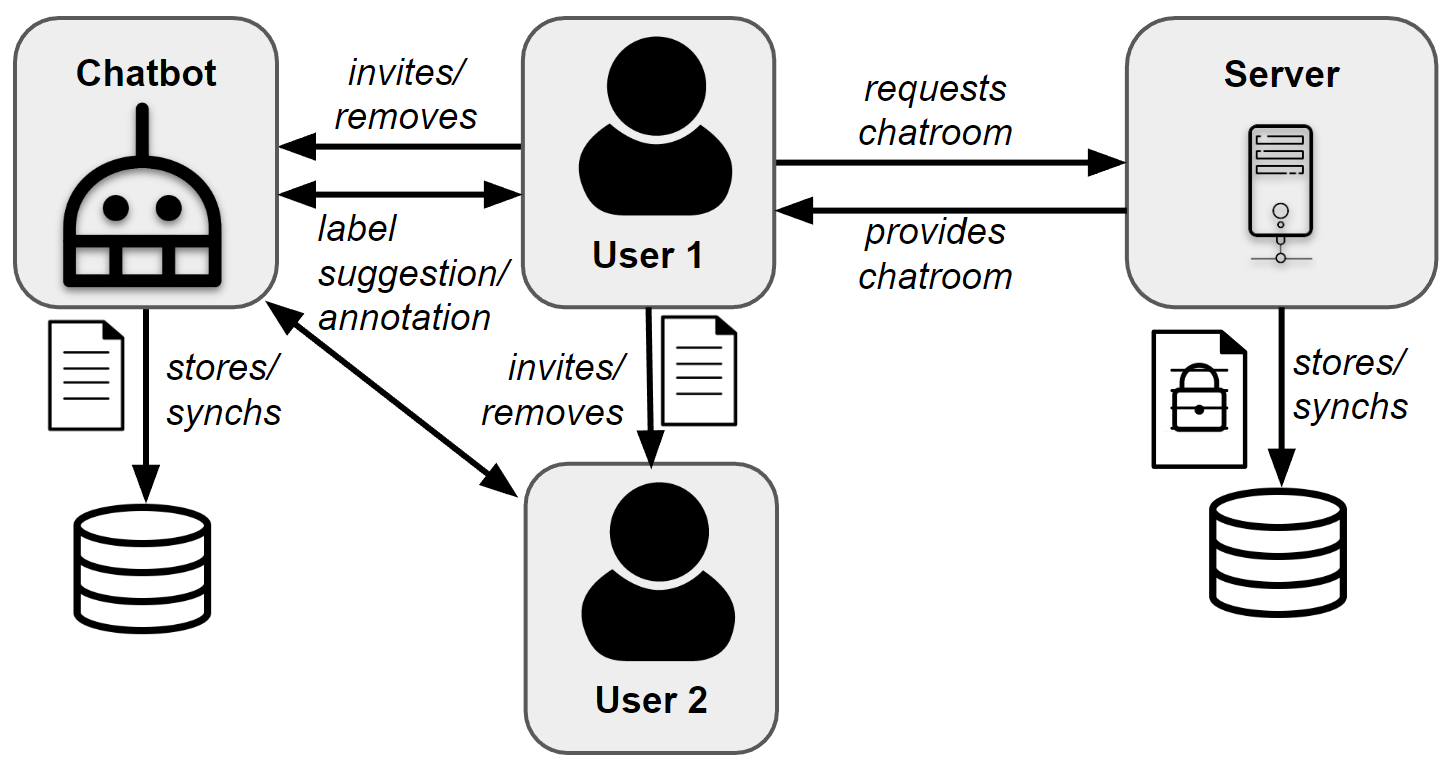}
    \caption{Information flow and privileges between users, server, and chatbot. While staying in a chatroom, the chatbot can decrypt all messages and stores them locally. Messages passed via the server are always encrypted.}
    \label{fig:system}
\end{figure}

\begin{figure}
    \centering
    \includegraphics[width=0.48\textwidth]{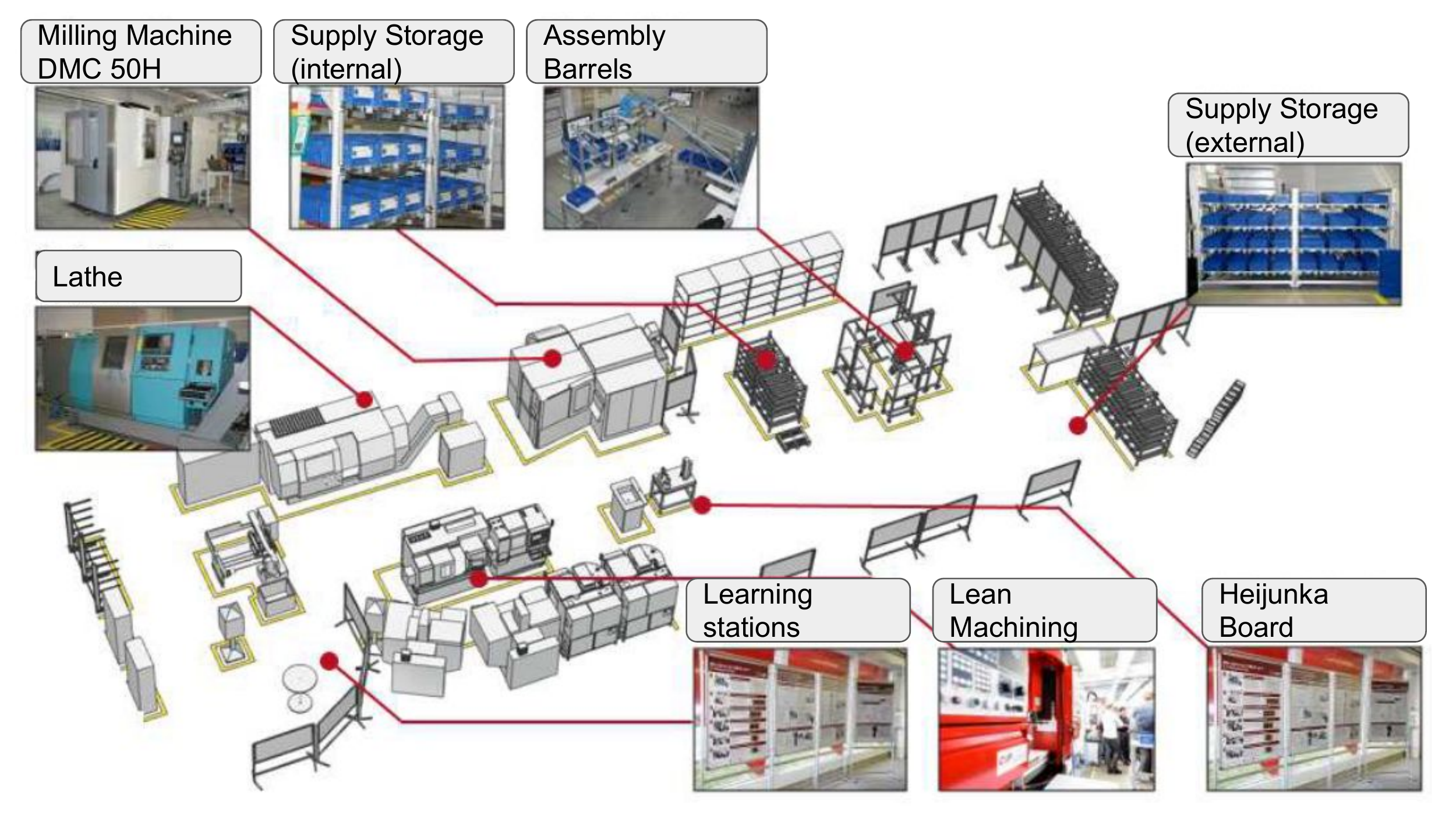}
    \caption{Workstations and machines in the Center for Industrial Productivity (CiP).}
    \label{fig:cip}
\end{figure}

\subsection{Production Environment}
Our working environment is the learning factory \textit{Center for Industrial Productivity} (CiP) at the Technical University of Darmstadt (TUDa).
The factory consists of various assembly stations, machines, and demonstrators (cf.~Figure~\ref{fig:cip}) and is run and maintained by $\sim$15 research assistants and 40 student assistants~\citep{muller2021extracting}.
One substantial challenge is the on-site support in case of problems that occur during their daily work, for instance, if a machine suddenly stops working. 
Usually, workers then ask their co-workers, technical support support, or the supervising research assistant (who may not be present) for assistance, often via informal communication channels.
While this leads to a quick fix of the issue, the knowledge of how to resolve such errors is not explicitly stored and hence, can be forgotten or lost over time.

\subsection{Preliminary Survey}
To assess the need of an NLP-based assistance system, we rely upon the analysis from a previous survey that was conducted at the CiP~\citep{tubiblio125537}. 
In this survey, they identify eight key issues and challenges from an employee's perspective.
(1) The most frequently used communication channel are emails.
(2) Most questions are answered fast, but in case of a slow return-rate it takes very long to receive an answer which leads to a substantial delay of the assembly line.
(3) There are no platforms that pool already encountered problems and solutions. Thus, there is a high demand for such a system.
(4) Most employees would use such an application only for work communication.
(5) A majority of employees are convinced that such an application could help in substantially reducing the required time to find a solution.
(6) Most employees are fine with using such an application on their private phone.
(7) All employees agreed to have a chatbot in a group chat monitoring the chatroom, but most stated that this would influence their communication behavior. 
(8) The most important benefit would be the improvement of knowledge management.

For companies, they identify three important criteria.
(1) A high level of data security is essential to avoid any leakage of information outside the company (i.e., the application should be self-hostable).
(2) No personal data may be processed to avoid legal complications.
(3) The most important benefit would be the improvement of error-reporting and -monitoring processes.

\section{System Description}\label{sec:approach}
As shown in Figure~\ref{fig:bot_user}, TexPrax involves three key parties: the users (employees), the chatbot and the server (e.g., hosted by a company).
Users communicate via chatrooms; each chatroom including at least two (for a private conversation) or more (for a group conversation) users.
Every message a user sends into a room can be read by any other user in the same room.
The server is responsible for handling new incoming messages and the distribution of outgoing messages, as well as keeping track of currently active conversations and users.
Finally, the chatbot is responsible for monitoring conversations, suggesting labels, and storing the relevant data (locally or in an external database). 

\begin{figure}
    \centering
    \includegraphics[width=0.48\textwidth]{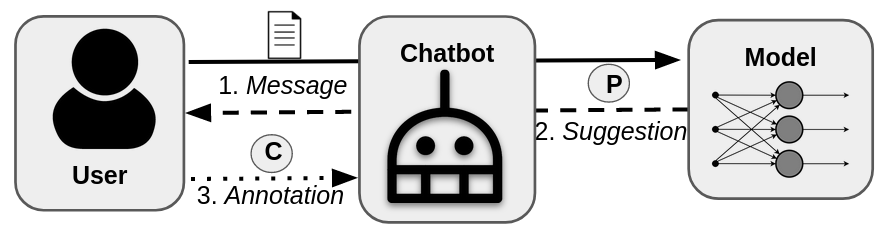}
    \caption{Information flow between the user, chatbot, and the underlying model.}
    \label{fig:bot_model}
\end{figure}

\subsection{Interaction and Privileges}
\label{sec:interaction}
A key focus of TexPrax lies within giving users full control over their data and when their conversation should be monitored. 
We thus provide them with the option to remove the chatbot from a conversation at any time.
Moreover, the matrix communication protocol allows users to modify and remove their messages which are then propagated to other participants in a chatroom including the chatbot.
This provides a safer communication space to users as they have full control over what messages are stored.
To comply with GDPR regulations~\citep{GDPR}, we further implement a feature to obtain the informed consent of the users for each chatroom. 

\begin{figure*}[!htb]
    \begin{subfigure}[b]{0.5\textwidth}
        \centering
        \includegraphics[width=1\textwidth]{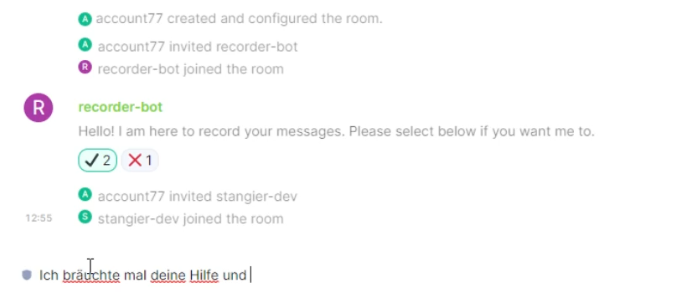}
        \caption{Introductory message after a new room has been created. If at least one user is against recording, the bot will leave the room.}
        \label{fig:new_room}
    \end{subfigure}
    \hfill 
    \begin{subfigure}[b]{0.48\textwidth}
        \centering
        \includegraphics[width=1\textwidth]{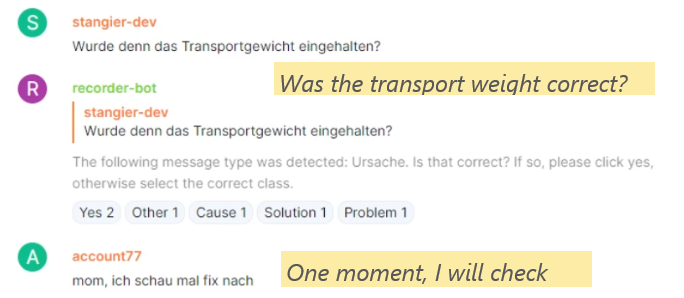}
        \caption{Label suggestion for a recognized cause.}
        %\vspace{0.5em}
        \label{fig:msg_classification}
    \end{subfigure}
    \caption{Example messages of TexPrax.}
\end{figure*}

Upon being invited to a chatroom, the chatbot automatically sends an introductory message and explicitly asks if this room shall be recorded (Figure~\ref{fig:new_room}).
The user can then respond to the question with one of the provided reactions.
If no reaction is selected but a message is sent, the chatbot will assume that the invitation was not intended and leaves the chatroom automatically without recording any message.
Only upon acceptance, the chatbot will notify the user and start monitoring and reacting to new messages (Figure~\ref{fig:msg_classification}).
Note, that the chatbot can be removed by the user at any time and invited back in later.
Due to the end-to-end encryption, the chatbot will not be able to read any messages that have been sent while it was not present. 

Annotations are then made by the user by accepting the suggested label, or providing a correction.\footnote{We also investigated labeling messages using free-text replies; however, users asked for an easier way of interaction.}
Only when a reaction is provided, the message and its class are stored in the internal database.

\subsection{Server}
The server is based on the Synapse implementation of the matrix protocol\footnote{\url{https://matrix.org/docs/projects/server/synapse}}; an open-source privacy-centric messaging protocol that enables end-to-end encrypted communication while allowing the server to be hosted on custom hardware~\citep{ermoshina2016end}.
This guarantees that all messages that are passed between users (and the chatbot) remain encrypted on the server and thus, cannot be read even if the server is breached.
The usage of Synapse further allows users to use different client applications such as Element\footnote{\url{https://element.io/}} across different platforms (i.e., mobile, desktop, and browser) to send and receive messages.
For the study and debugging purposes, we further extended the existing implementation to automatically send an invitation to the chatbot every time a new chatroom is created (users will still be asked for their consent before recording any messages).
TexPrax is setup on a virtual machine with 4 CPU cores, 8 GB RAM, and 50 GB of storage.

\subsection{Chatbot}
The chatbot is based on the nio project\footnote{\url{https://matrix.org/docs/projects/sdk/matrix-nio}}---a client library for the matrix protocol---written in Python.
As soon as the user allows the chatbot to record messages, it will store every new message including the annotation into a local database.
Processing messages can freely be extended; for instance, it is also possible to send the messages to an external instance via HTTP instead storing them locally.
To provide the system with additional flexibility, the chatbot can be hosted completely separate from the server.
It is thus possible to run different chatbots for each chatroom on different hardware, which can be helpful to better comply with data privacy regulations.
As shown in Figure~\ref{fig:bot_model}, we utilize a pre-trained model to provide users with label suggestions. 
The chatbot will then react to a message with a label suggestion and ask the user to confirm or correct the notification (they can also just ignore the message).
All user annotations are stored separately from the model's suggestion.

\section{Data Collection}\label{sec:data_collection}
\begin{table*}[!htb]
    \centering
    \resizebox{0.8\textwidth}{!}{
    \begin{tabular}{@{}crcr|rrrrr|c@{}} 
    \toprule
        Part &  Dialogues (D) & Turns (T) & T/D & Problem & Solution & Cause & Other & Total & Sents/D\\ \midrule
        P1 &  81 & 246 & 3.04 & 127 & 74 & 50 & 302 & 553 & 6.83 $\pm$ 3.82 \\ 
        P2 & 97 & 309 & 3.19 & 117 & 56 & 114 & 145 & 432 & 4.45 $\pm$ 2.11 \\
        P3 & 24 & 36  & 1.50    & 23 & 12 & 1 & 6 & 42 & 1.75 $\pm$ 0.66 \\
        \midrule 
        \textbf{Total} & 202 & 591 & 2.60 & 267 & 142 & 165 & 453 & 1,027 & 5.08 $\pm$ 3.28 \\
        \bottomrule
    \end{tabular}
    }
    \caption{The number of dialogues, turns, and their ratio (left) and the class distribution on a sentence-level (right).}
    \label{tab:dataprops}
\end{table*}

In contrast to our previous work that investigates expert-annotated named entity recognition~\citep{muller2021extracting}, our goal is to provide a first solution for collecting annotated data and providing assistance with a minimal effort for users.
We thus focus on sentence-level annotations that can be easily provided using message reactions and are suitable for existing shop floor management processes.

\subsection{Annotation Task}
Following existing workflows for shop floor management that are currently done on paper, we identify three crucial classes for our use case:

\begin{enumerate}
    \item Problem (\textbf{P}): The description of a deviation from an expected target state, e.g., machine breakdowns, material delays, incorrect production processes etc. (often formulated as a question). 
    \item Cause (\textbf{C}): The assumed cause of a problem.
    \item Solution (\textbf{S}): The action to eliminate the root cause of the problem or to help in finding the possible causes and countermeasures.
    \item Other (\textbf{O}): None of the above classes (e.g., unrelated messages).
\end{enumerate}

To train an initial label suggestion model, we re-annotated the existing dataset with sentence-level annotations.\footnote{Deploying TexPrax without any suggestion model does not affect the number of reactions provided by users.}
This was done by three of the authors that are responsible for managing the CiP.
Each of them annotated one third of the dataset and cross-examined all other annotations for possible errors or disagreement.
Upon disagreeing on a label, all annotators discussed the respective instance to agree upon the best suited one.

\subsection{Participants}
All participants were student assistants, technical support staff, or researchers that worked in the CiP and are employed at the university; receiving payment according to the official wages (above German minimum wage).
They were informed about the purpose of the study in advance, and provided their informed consent before participation.
They further received instructions about how to use the application including the features allowing them to modify and remove already sent messages. 
Participation was strictly voluntary and anonymous; to further obfuscate the identity of our participants, we created a pool of user accounts from which an account was randomly assigned to each user.
For data publication we obfuscate the user accounts by hashing the ID of each user.
Overall, our study had a total number of 10 participants over the whole duration (October 2021 to July 2022).

\subsection{Data Analysis}
Table~\ref{tab:dataprops} shows the statistics of the collected data.
We split the data into three parts; first, the re-annotated dataset that was used to train the label suggestion model (cf. Section~\ref{sec:evaluation_study}), second, data collected between October 2021 and June 2022, and third, the data collected in July 2022 to evaluate the our final system which we also use as the test data for our experiments.
The second and third batch of data was each collected in a separate chatroom.
An overview of the dialogue properties can be found in Table~\ref{tab:dataprops}.
Overall, the dataset consists of 202 dialogues with 591 turns and 1,027 sentences. 
A close inspection of the data reveals interesting properties (e.g., grammatically incorrect language, abbreviations, etc.).
Despite that, we want to emphasize that there was no single case where our participants could not understand a message.

\paragraph{Distributional shifts.}
Table~\ref{tab:dataprops} shows varying class distributions across all three splits.
One reason for this may be the amount of expertise in chatrooms across different periods of data collection.
For instance, between the first and second part of the data collection which were $\sim$10 months apart, there had been a partial change of staff in work force.
With new people joining the CiP, we find a higher number of responses looking for potential causes of a problem, but with less success (i.e., less solutions).
We further find that the more acquainted workers in the first data collection tend to provide longer explanations and engage themselves more in chitchat which is reflected in the substantially higher number of \textit{Other} class sentences and a higher sentence-per-dialog ratio (Sents/D).

\paragraph{Slang.} We find various occurrences of text mimicking spoken language involving grammatically incorrect expressions.
For instance, our participants frequently used \textit{ne} instead of \textit{eine} (\textit{Eng.:} a/an) or as the short form of \textit{nein} (\textit{Eng.:} no).

\paragraph{Abbreviations.}
We find that our participants tend to communicate in short messages that involve abbreviations.
While some are easily understandable for native German speakers---e.g., \textit{vllt.} for \textit{vielleicht} (\textit{Eng.:} maybe)---others such as \textit{V8} for \textit{Variante 8} (a product type) or \textit{wimi} for \textit{wissenschaftliche Mitarbeitende} (\textit{Eng.:} researcher) are highly dependent on the domain.

\paragraph{Filler words.} Similar to in-person conversations, we also find an abundance of filler words such as \textit{ah}, \textit{hmm}, and \textit{oh}. 

\paragraph{Code switching.} We find that participants sometimes tend to code switch from German to English~\citep{scotton1977bilingual}; especially for short, one word responses (e.g., \textit{Nice!}, \textit{Sorry!}).

\section{Experiments}\label{sec:evaluation_study}
We conduct experiments to gain first insights on how well recent models can perform for providing label suggestions for our use case in future studies.

\subsection{Experimental Setup}

We evaluate two models that are capable of processing German texts as our baselines. 
First, the XLMR-base model~\citep{conneau2019unsupervised} provided by Huggingface~\citep{wolf2019huggingface} that has been shown to have a solid performance across various languages~\citep{malmasi-etal-2022-semeval}.
Second, a German version of BERT (GBERT, \citealt{chan-etal-2020-germans}).
This has been shown to work well for German tweets that have a similar format (i.e., short, German sentences containing informal language) as our messages~\citep{beck-etal-2021-investigating}.
For sentence classification, we use the $\mathrm{[CLS]}$ token to predict if a given sentence states a \textit{problem} (P), a \textit{cause} (C), a \textit{solution} (S), or \textit{other} (O).
Across all experiments, we train our models for $10$ epochs with a learning rate of  $2 \mathrm{e}^{-5}$ and weight decay of $0.01$, and a batch size of $16$.
We use the parts 1 and 2 as presented in Table~\ref{tab:dataprops} for training and use part 3 as the most recently collected dataset for testing.

\begin{table}[!htb]
\centering
\resizebox{0.98\columnwidth}{!}{
\begin{tabular}{@{}lcccccc@{}} \toprule
  &  \multicolumn{2}{c}{P1} & \multicolumn{2}{c}{P2} & \multicolumn{2}{c}{P1 + P2} \\ \midrule
 Model & Acc & F1 & Acc & F1 & Acc & F1 \\ \midrule
XLMR & 0.357 & 0.216 & 0.524 & 0.315 & 0.476 & 0.269   \\
GBERT & 0.405 & 0.267 & 0.310 & 0.237 & 0.429 & \textbf{0.361} \\ \bottomrule
\end{tabular}
}
\caption{Accuracy and macro-averaged F1 scores of both models trained on different temporal datasplits.}
\label{tab:results}
\end{table}

\subsection{Results}

Table~\ref{tab:results} shows the results of both models on the P3 data (cf. Table~\ref{tab:dataprops}).
Both models are not able to achieve a marco-averaged F1 score higher than $0.4$, showing that even recent language-specific models struggle for sentence classification when applied to a very specific domain and little training data (432--553 sentences).
Interestingly, GBERT outperforms XLMR when trained on P1 data as well and when trained on P1 $+$ P2 data in terms of F1 score.
Although we initially conjectured that XLMR may be capable of better handling the code switched data, this does not seem to be the case. 
We further find that the suggested labels from the GBERT model (trained on P1 data) during the collection of P2 achieved an accuracy of 0.683.
While this is a moderately high performance, this also implies that 31.7\% of the labels needed to be corrected by our participants. 

\subsection{Usability}
To ensure that this did not substantially impact the usability of TexPrax, we asked our voluntary participants to answer the system usability scale (SUS) questionnaire~\citep{brooke1996sus} upon finishing the final round of data collections (P3).
SUS quantifies the relative usability with respect to existing benchmarks and ranges from \textbf{A$^+$} (84.1--100 SUS) to \textbf{F} (0--51.6 SUS)~\citep{lewis2018item}.
Overall, seven users participated in the usability study.
On average, TexPrax receives a system usability scale score of $81.76$ with a standard deviation of $5.46$, which indicates an \textbf{A} level (80.8--84.0 SUS) usability. 
We thus conclude that TexPrax achieves a high usability despite the label corrections.

\section{Conclusion}\label{sec:conclusion}
We presented TexPrax, a system for collecting annotations and assisting employees by directly engaging them as domain-experts during their daily work.
TexPrax allows users to exchange, modify, and delete end-to-end encrypted messages at any time, and an opt-in chatbot to ensure a high level of data privacy and security.
We evaluate TexPrax in an assembly line at a learning factory (CiP) where we find that daily work communication is noisy, but efficient and very problem-oriented.
While existing models still have difficulties to provide the correct label suggestion, TexPrax still maintains a high usability.
We conjecture that TexPrax could be especially beneficial to collect data and build assistance systems in domains with a high share of remote work, such as in software development.
For future work, we plan to extend TexPrax to identify and suggest solutions for recognized problems and adapt it to new domains, such as our institute's reading group chat where researchers discuss papers relevant for their research.

\section*{Acknowledgements}
We thank all subjects from the user study and our anonymous reviewers, Max Glockner, and Jan-Christoph Klie for their helpful feedback.
This work has been funded by the European Regional Development Fund (ERDF) and the Hessian State Chancellery – Hessian Minister of Digital Strategy and Development under the promotional reference 20005482 (TexPrax).

\section*{Ethics Statement}
The collection of data from group- and private-chats requires careful consideration about what kind of data is to be expected and how users can control it.
To ensure an ethical data collection and usage, we worked closely together with the respective bodies of our university (TUDa) for developing our final workflow.
We want to emphasize that such data should never be collected without the explicit and informed consent of the users.
Our participants voluntarily participated in this study and furthermore, had an active interest in the system as they could directly benefit from it.

\paragraph{Pre-study clearance from respective bodies.}
After defining our data collection workflow and annotation task, we hence asked the ethics committee of our university for ethical clearance.\footnote{\url{https://www.intern.tu-darmstadt.de/gremien/ethikkommisson/index.en.jsp}}
To further ensure the (mental) safety of our participants who were employees of TU Darmstadt, we further asked our university's staff council for their clearance.\footnote{\url{https://www.personalrat.tu-darmstadt.de/personalrat_1/index.de.jsp}} 
Both bodies provided their full clearance to conduct this study after minor modifications of the initial workflow involving the account distribution to participants (cf. Section~4.2).
Both clearance letters for the final study setup can be shared upon request (in German).

\paragraph{Informed consent.}
All our participants were fully informed about the data collection processes, for what purpose the data was collected, and how it will be used and released (including the surveys). 
They all provided their informed consent before requesting an anonymous user account for participation in the study (this was a mandatory requirement from the ethics committee and staff council).

\section*{Limitations}

\paragraph{Interactive assistance.} 
In this work, we focused on data collection and annotation from workers in a factory environment. 
Although the integration of TexPrax into their existing dashboard\footnote{\url{https://www.sfmsystems.de/}} alleviates their daily work, additional assistance could be provided by automatically suggesting solutions for identified problems.

\paragraph{Other use cases.} While TexPrax received clearance by our university's ethics committee and staff council, it must be noted that this does not automatically transfer to new use cases or even similar ones at different universities/factories.
It is crucial to get at least clearance of the respective staff council before deploying TexPrax to avoid any legal issues that may otherwise arise.
Moreover, for the collected data to be of use for the NLP community, the company (or a respective organization) must be willing to share their data publicly. 
This however implies that deploying TexPrax in organizations that handle sensitive data (e.g., security-related or personal user data) can alleviate the work of employees, but will not result in datasets that can be publicly shared. 

\paragraph{Different annotation tasks.} The current version of TexPrax is designed as a tool for collecting data and annotations on a sentence-level.
Explicitly asking for free-text responses could be one solution to tackle different kinds of annotations such as identifying named entities---for instance, a user could reply to a message containing a named entity by repeating it---however, this may hurt usability and lead to a less frequent usage of the application.
To extend TexPrax to different annotation tasks one thus first needs to find a good way to interact with the user.

\paragraph{Propagating dataset changes in trained models.}
Finally, a last limitation is updating the training data that is implicitly stored in the trained model. 
The lack of efficient methods to update only specific information in trained models can lead to a substantial overhead when implementing changes in the data made by a user as the whole model needs to be retrained.

\bibliography{bibliography}
\bibliographystyle{acl_natbib}

\appendix

\end{document}